\documentclass[10pt,journal,comsoc]{IEEEtran}
\ifCLASSOPTIONcompsoc
  \usepackage[nocompress]{cite}
\else
  \usepackage{cite}
\fi

\usepackage{graphicx}

\usepackage{multirow,multicol}

\usepackage{amsmath}
\usepackage[cmintegrals]{newtxmath}

\usepackage{array}

\ifCLASSOPTIONcompsoc
 \usepackage[caption=false,font=footnotesize,labelfont=sf,textfont=sf]{subfig}
\else
 \usepackage[caption=false,font=footnotesize]{subfig}
\fi

\usepackage{url}

\begin{document}

\title{Cross-Task Consistency Learning Framework for Multi-Task Learning}

\author{Akihiro~Nakano, Shi~Chen, and~Kazuyuki~Demachi
\thanks{A. Nakano is with Department of Systems Innovation, Faculty of Engineering, The University of Tokyo, Tokyo, Japan (e-mail: nakano.akihiro@weblab.t.u-tokyo.ac.jp).}
\thanks{S. Chen, and K.Demachi are with Department of Nuclear Engineering and Management, School of Engineering, The University of Tokyo, Tokyo, Japan (e-mail: shichen@g.ecc.u-tokyo.ac.jp; demachi@n.t.u-tokyo.ac.jp)}}
\maketitle

\begin{abstract}
Multi-task learning (MTL) is an active field in deep learning in which we train a model to jointly learn multiple tasks by exploiting relationships between the tasks. It has been shown that MTL helps the model share the learned features between tasks and enhance predictions compared to when learning each task independently. We propose a new learning framework for 2-task MTL problem that uses the predictions of one task as inputs to another network to predict the other task. We define two new loss terms inspired by cycle-consistency loss and contrastive learning, \emph{alignment loss} and \emph{cross-task consistency loss}. Both losses are designed to enforce the model to align the predictions of multiple tasks so that the model predicts consistently. We theoretically prove that both losses help the model learn more efficiently and that cross-task consistency loss is better in terms of alignment with the straight-forward predictions. Experimental results also show that our proposed model achieves significant performance on the benchmark Cityscapes and NYU dataset.
\end{abstract}



%
\IEEEpeerreviewmaketitle

\ifCLASSOPTIONcompsoc
\IEEEraisesectionheading{\section{Introduction}\label{sec:introduction}}
\else
\section{Introduction}\label{sec:introduction}
\fi
\IEEEPARstart{D}{eep} learning has made significant progress in the last decade, improving and enhancing its ability to classify, predict, and understand inputs from various modals. Although its performance are now excelling the skills of humans in certain domains, one disadvantage of deep learning models is that its application is limited to single-task problems. While humans are capable of handling multiple tasks simultaneously, common deep learning models require different models to be trained for each task. Therefore, recent works have focused on developing a multi-task learning model, a model optimized for multiple tasks.

Multi-Task Learning (MTL) is a method to train a model to learn multiple tasks jointly. Humans are able to process multiple tasks at the same time and combine the information for a more semantic task. For example, when we look at an image, we are able to recognize objects, estimate depth, infer the scene, and predict what will happen next. One reason we can process and integrate multiple sensory tasks is because they are closely related to each other. MTL aims to exploit this ``relationship'' in a way such that machines can utilize it. 

Recent MTL works have investigated sharing full or part of the architecture. Some works have proposed a method to explicitly enforce the model to capture relationships between tasks using additional units or matrices \cite{long17,misra16,yang19}. Other works have simply divided the features into task-common features and task-specific features so that it learns the relationships implicitly and focused on balancing the losses between multiple tasks \cite{liu19,jha20,kendall18,chen18,yu20}.

Less research has been conducted for self-supervised learning of MTL. Works such as \cite{chen19,klingner20} have utilized the stereo RGB image pairs for better semantic consistency or better depth estimation. However these methods require the data to be collected using stereo cameras and therefore cannot be applied if the data were collected otherwise (ex. NYU dataset \cite{silberman12} which was collected using Microsoft Kinect camera, an infrarerd camera). To the best of our knowledge, no prior works have explored a self-supervised MTL which can be used for both stereo cameras and infrared camera.


In this work, we define and propose a new framework called \textit{cross-task consistency learning framework} for 2-task MTL problem inspired by CycleGAN's \cite{zhu17} cycle-consistency loss and contrastive learning using multi-views \cite{chen20}, \cite{tosh20}. We introduce two loss terms, \textit{alignment loss} and \textit{cross-task consistency loss}, aimed to maintain consistency between the predictions. We then prove an inequality relationship between alignment loss and cross-task consistency loss and show that not only cross-task consistency loss is superior to alignment loss in approximating the straight-forward predictions but also that both terms are upper-bounded by a small value.

Specifically, we demonstrate our method in learning semantic segmentation and depth estimation task in the computer vision field. We use these two tasks because (i) scene segmentation and depth perception are related to each other \cite{burge10}, and (ii) the encoder-decoder architecture which has shown significant performance for both tasks is suited to capture task-common and task-specific features. Our architecture, which we name as XTasC-Net (Cross-Task Consistency Network), consists of two modules: first, the input image is feeded to an encoder-decoder architecture to produce predictions of the two tasks (direct predictions). Then, we prepare two separate encoder-decoder architecture networks (Task-Transfer Networks, TTNets) that takes in the direct predictions as inputs and outputs the prediction for the other task (task-transferred predictions). The proposed alignment loss and cross-task consistency loss utilizes the task-transferred predictions so that some of the information from the other task is learned through the scope of the TTNets. The experimental results on Cityscapes \cite{cordts16} and NYU \cite{silberman12} dataset shows that our model shows competitive performance with less number of parameters.

In summary, we make mainly two contributions. First, we present a principled multi-task learning framework for 2-task MTL problem by relating to cycle-consistency and contrastive learning. We theoretically derive our loss term using implicit latent variable models (LVMs) and prove an inequality relationship regarding the proposed two loss terms. Secondly, we propose a new architecture called XTasC-Net for semantic segmentation and depth estimation task. Our method is not only applicable to data collected via stereo cameras but also monocular cameras.

\section{Related Work}\label{sec:related_work}
MTL has been one of the key approaches in improving generalization and learning efficiency by using information of \textit{related} tasks \cite{caruana98,ruder17}. A common challenge of MTL is the formulation of task relationships. Some works have approached using regularization methods \cite{micchelli04} or by defining a convex optimization problem to estimate task relationships \cite{ciliberto10,zhang10,ciliberto17}. Zamir \textit{et al.} \cite{zamir18} built a taxonomical structure between multiple vision-related tasks. Furthermore, Zhou \textit{et al.} \cite{zhou21} used adversarial networks to learn the task relationships.

In the computer vision field, some works explored modeling task relationships explicitly in their model architecture. For example, Long \textit{et al.} \cite{long17} developed Deep Relationship Networks which places a matrix prior between the fully-connected layers of each task so that the model learns the task relationship. Misra \textit{et al.} \cite{misra16} proposed Cross-stitch Networks which uses cross-stitch units, a module that learns how much sharing is needed between each respective layer of the networks. Yang \textit{et al.} \cite{yang19} used min-max optimization to make the model learn both task-specific and task-common features.

On the contrary, other works have experimented enforcing the model to learn task relationships implicitly. There have been mainly three approaches: using knowledge distillation, balancing multi-task loss function, and utilizing stereo views of the data.

One approach has been to use knowledge distillation \cite{hinton15} by preparing two phases of training to enhance performance \cite{xu18,vande19,zhang19,li20,vande20}. These models are trained on several tasks jointly in the first phase. Then, using the trained weights from the first phase, the features are combined through another network to distill its information to make the final predictions. For example, Xu \textit{et al.} \cite{xu18} proposed PAD-Net which experiments concatenation of the features or applying an attention map when passing features of other tasks for a certain task. Zhang \textit{et al.} \cite{zhang19} and Vandenhende \textit{et al.} \cite{vande20} utilized self-attention modules in each task's network. Li et al.'s \cite{li20} model used knowledge distillation by adding a loss term between the pretrained STL network and MTL model's weights.

Another approach is to train a single model by balancing the loss functions of each task \cite{mousa16,chen18,kendall18,zhang18,sener18,liu19,jha20,yu20}. Kendall \textit{et al.} \cite{kendall18} proposed uncertainty weights, a quantity that can be seen as the relative confidence between tasks. Other methods \cite{chen18,jha20} have used loss magnitude to leverage the losses. On the other hand, Liu \textit{et al.} \cite{liu19} introduced DWA (dynamic weight averaging) which uses relative loss reduction and Yu \textit{et al.} \cite{yu20} implemented PCGrad, an algorithm that fixes contradicting gradients in the shared architecture.

Finally, some works for MTL including depth estimation task has utilized the multiple views of the data \cite{chen19,klingner20}. Using the advantage that some dataset were captured using a stereo camera, these methods have applied the photometric reconstruction loss originally proposed by MonoDepth \cite{godard17,godard19}. 

\section{Method}\label{sec:method}
We use the following notations: a common network architecture in recent works consists of two modules, a shared network $f_{W_E}:\mathcal{X}\rightarrow\mathcal{\tilde{X}}$ followed by $T$ individual networks $f_{W_t}:\mathcal{\tilde{X}}\rightarrow\mathbb{R}^{d_t}$ for each task where $W_E, W_t$ are the weights of each networks for $t\in[T]$. Since the number of tasks is limited to $T=2$, we refer the two tasks as $y$ and $z$. Further, for simplicity, we regard the $\tilde{x}=f_{W_E}(x)$ as our input and denote this as $x$.

Our key assumption is that there exists some mappings $\mathcal{F}_\theta:\mathcal{Z}\rightarrow\mathcal{Y}$ and $\mathcal{G}_\phi:\mathcal{Y}\rightarrow\mathcal{Z}$ that describes the likelihood between the tasks,
\begin{equation}
\begin{aligned}
y&\sim p_\theta(y|z) \Leftrightarrow y=\mathcal{F}_\theta(\varepsilon;z),&\;\;&\varepsilon\sim p(\varepsilon)\\
z&\sim p_\phi(z|y) \Leftrightarrow z=\mathcal{G}_\phi(\varepsilon^\prime;y),&\;\;&\varepsilon^\prime\sim p(\varepsilon^\prime)\label{eq:mapping_assumption}
\end{aligned}
\end{equation}
where $\varepsilon,\varepsilon^\prime$ are noise.

Since $y$ and $z$ are both outputs from a neural network given an input $x$, we can further denote this using $f_{W_t}$ as,
\begin{equation}
\begin{aligned}
y&=\mathcal{F}_\theta(\varepsilon;f_{W_2}(x))\\
z&=\mathcal{G}_\phi(\varepsilon^\prime;f_{W_1}(x))
\end{aligned}
\end{equation}
Using this notation, we propose \emph{cross-task consistency loss}, which is defined as,
\begin{equation}
\begin{aligned}
\ell_{2\rightarrow1}^{\rm XTC}&=\|\mathcal{F}_\theta(f_{W_2}(x))-f_{W_1}(x)\|_2^2\\
\ell_{1\rightarrow2}^{\rm XTC}&=\|\mathcal{G}_\phi(f_{W_1}(x))-f_{W_2}(x)\|_2^2
\label{eq:xtc}
\end{aligned}
\end{equation}
where $\|\cdot\|_2$ indicates $l_2$ norm. The loss takes the difference between the outputs of task-specific networks, $f_{W_1}(x),f_{W_2}(x)$, and those outputs that are transferred to predict the other task, $\mathcal{G}_\phi(f_{W_1}(x)),\mathcal{F}_\theta(f_{W_2}(x))$. Below, we refer the former as \emph{direct predictions} and the latter as \emph{task-transferred predictions}.

The intuition of cross-task consistency loss is to pass partial information of task-specific features to other tasks using the assumed mappings, $\mathcal{F}_\theta,\mathcal{G}_\phi$. Although the shared network helps the model learn these task-common features, we expect some of the task-specific features learned in each task-specific network can also be utilized for the other task.

In the following sections, we derive our proposed loss term from \cite{tiao18,tosh20}. First, in section \ref{sec:align_loss}, we derive a similar loss term which we name as alignment loss from Tiao \textit{et al.}'s \cite{tiao18} proof of CycleGAN based on implicit latent variable models (LVMs). Then, in section \ref{sec:xtc}, we will prove that cross-task consistency loss is better than alignment loss based on Tosh et al.'s proof \cite{tosh20}.

\subsection{Alignment Loss}\label{sec:align_loss}

\begin{figure}[!t]
\centering
\includegraphics[width=3.3in]{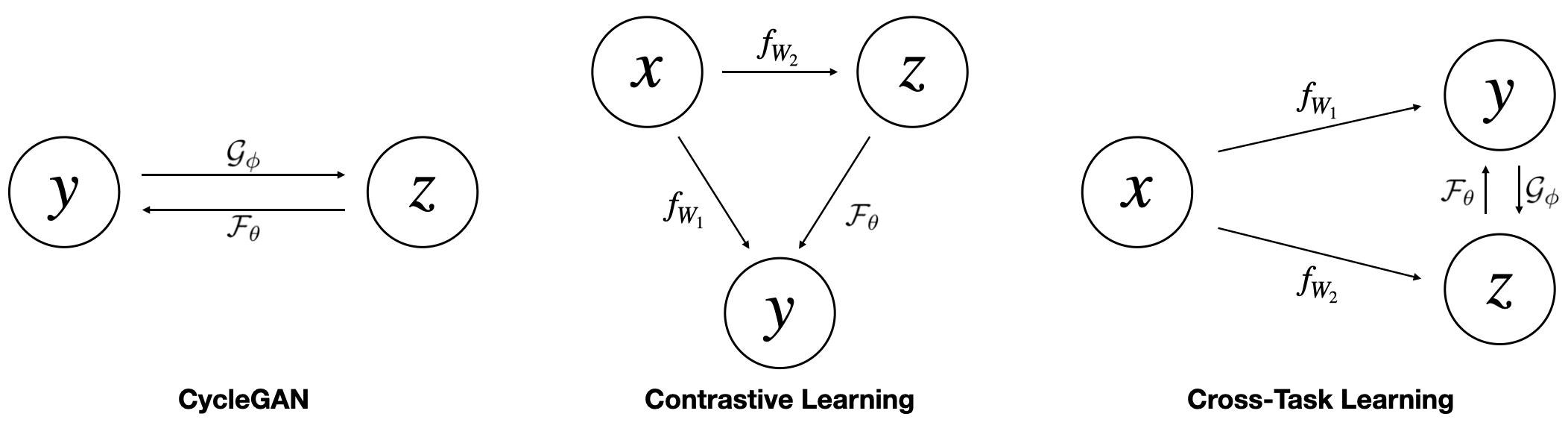}
\caption{Diagrams of CycleGAN (left), contrastive learning (center), and our cross-task consistency learning framework (right).}
\label{fig:cycle_cl_xtc}
\end{figure}

Fig. \ref{fig:cycle_cl_xtc} describes the diagrams of CycleGAN and our cross-task consistency learning framework. Similar to how CycleGAN has a bidirectional mapping between the input $y$ (real images) and the output $z$ (generated images), our framework also a bidirectional mapping between the two tasks' outputs.

Using LVMs, we can describe the joint probability of observing $y$ and $z$, the two tasks, as the product of the prior distribution and the likelihood. Since we model the likelihoods using  \eqref{eq:mapping_assumption}, we can express the joint probability as,
\begin{equation} 
\begin{aligned} \label{eq:lvm}
p_\theta(y,z)&=p_\theta(y|z)p(z)\\
q_\phi(y,z)&=q_\phi(z|y)q(y)
\end{aligned}
\end{equation}
where $p(\cdot),q(\cdot)$ denotes probability. We further use Tial \textit{et al.}'s implicit LVMs, in which we replace the prior distributions with implicit distributions $p^*(u)$. Implicit distributions are distributions given by a finite collection of data, $U^*=\{u_i^*\}_{i=1}^N$, i.e. $u_i^*\sim p^*(u)$, for $u\in\{y,z\}$. Therefore, \eqref{eq:lvm} can be written as,
\begin{equation} 
\begin{aligned} \label{eq:implicit_lvm}
p_\theta(y,z)&=p_\theta(y|z)p^*(z)\propto p_\theta(y|z)p^*(x)\\
q_\phi(y,z)&=q_\phi(z|y)q^*(y)\propto q_\phi(z|y)q^*(x)\\
\end{aligned}
\end{equation}
since both $y$ and $z$ are conditioned on input $x$.

Now, regardless of using parameters $\theta$ or $\phi$, the joint distribution should be equivalent. Therefore, we consider minimizing the statistical distance between $p_\theta(y,z)$ and $q_\phi(y,z)$ using \textit{symmetric} KL divergence, ${\rm KL}_{\rm SYMM}[p_\theta(y,z)\|q_\phi(y,z)]$, where
\begin{equation}
{\rm KL}_{\rm SYMM}[p\|q]={\rm KL}[p\|q]+{\rm KL}[q\|p]
\end{equation}
Then, since minimizing the KL divergence is equivalent to maximizing the likelihood in  \eqref{eq:implicit_lvm}, we can derive the loss as follows;

Assume Gaussian noise, i.e., $\varepsilon\sim\mathcal{N}(0,\sigma_1^2I)$, $\varepsilon^\prime\sim\mathcal{N}(0,\sigma_2^2I)$. Then,
\begin{equation*}
\begin{aligned}
y&\sim\mathcal{N}(\mathcal{F}_\theta(z),\sigma_1^2I)\\
z&\sim\mathcal{N}(\mathcal{G}_\phi(y),\sigma_2^2I)
\end{aligned}
\end{equation*}
Therefore using maximum likelihood estimation,
\begin{equation}
\begin{split}
\max p_\theta(y,z)&\iff\min\mathbb{E}_{p^*(z)p_\theta(y|z)}[-\log p_\theta(y|z)]\\
\therefore\;\;
\mathbb{E}&_{p^*(z)p_\theta(y|z)}[-\log p_\theta(y|z)]\\
&=\frac{1}{2\sigma_1^2}\mathbb{E}_{p^*(z)p_\theta(y|z)}[\|y-\mathcal{F}_\theta(z)\|_2^2]\\
&\:\:\:+\frac{D}{2}\log2\pi\sigma_1^2\\
&=\gamma_1\mathbb{E}_{p^*(z)p_\theta(y|z)}[\|y-\mathcal{F}_\theta(z)\|_2^2]+\delta_1\\
&\propto\mathbb{E}_{p^*(z)p_\theta(y|z)}[\|y-\mathcal{F}_\theta(z)\|_2^2]\\
&\propto\mathbb{E}_{p^*(x)p_\theta(y|z)}[\|y-\mathcal{F}_\theta(f_{W_2}(x))\|_2^2]
\end{split}
\end{equation}
where $\gamma_1=\frac{1}{2\sigma_1^2}$ and $\delta_1=\frac{D}{2}\log\frac{\pi}{\gamma_1}$. A similar thing can be said for maximizing $q_\phi(y,z)$.
Hence,
\begin{equation}
\begin{split}
\ell_{2\rightarrow1}^{\rm ALIGN}&=\|y-\mathcal{F}_\theta(f_{W_2}(x))\|_2^2\\
\ell_{1\rightarrow2}^{\rm ALIGN}&=\|z-\mathcal{G}_\phi(f_{W_1}(x))\|_2^2
\label{eq:alignment}
\end{split}
\end{equation}
are the \emph{alignment losses}. However, we introduce another loss, cross-task consistency loss, which is defined as \eqref{eq:xtc}, by replacing the first term of each equation with the direct predictions. In the following section, we explain the merits of using cross-task consistency loss over alignment loss.

\subsection{Cross-Task Consistency Loss}\label{sec:xtc}

Our learning framework can also be related with contrastive learning using multi-views (Fig. \ref{fig:cycle_cl_xtc}). Methods such as SimCLR \cite{chen20} learns in a self-supervised fashion by creating additional inputs that are correlated with the original input. Whereas contrastive learning uses the redundancy between the inputs $x$ and the generated images $z$, our framework has redundancy between the two tasks.

Below, we use the following notations to express direct predictors, alignment (ALIGN) predictors, and cross-task consistency (XTC) predictors,
\begin{equation}
\begin{aligned}
{\rm direct:}\;\;x&\mapsto\mathbb{E}[Y|X=x]\\
{\rm ALIGN:}\;\;x&\mapsto\mathbb{E}[Y|\mathbb{E}[Z|X=x]]\\
{\rm XTC:}\;\;x&\mapsto\mathbb{E}[\mathbb{E}[Y|X=x]|\mathbb{E}[Z|X=x]]
\end{aligned}
\end{equation}
\setlength{\arraycolsep}{5pt}
Furthermore, without proof, we can state that the quantity,
\begin{equation}
\xi_Y=\mathbb{E}[(Y-\mathbb{E}[Y|X])^2]
\end{equation}
is small.

By assuming that there exists some relationship between $y$ and $z$, predicting $y$ by first predicting $z$ from $x$ should, intuitively, achieve similar predictions as directly predicting $y$ from $x$.

The following proposition tells us that not only does this strategy work but also using the direct predictions as the target works better.

\textbf{Proposition.} \emph{Let X, Y, Z be random variables. Then, cross-task self-consistency loss yields smaller expected error between direct loss compared to cross-task consistency loss, i.e.,}
\setlength{\arraycolsep}{0.0em}
\begin{align*}
0&=\mathbb{E}[(\mathbb{E}[\mathbb{E}[Y|X]|\mathbb{E}[Z|X]]-\mathbb{E}[Y|X])^2]\\
&\leq\mathbb{E}[(\mathbb{E}[Y|\mathbb{E}[Z|X]]-\mathbb{E}[Y|X])^2]\\
&\leq\xi_Y
\end{align*}
\setlength{\arraycolsep}{5pt}
\emph{This holds true for case where $Y$ and $Z$ are switched.}

See Appendix \ref{sec:appendixA} for proof.

The proposition implies (1) cross-task consistency loss yields a smaller expected difference between direct loss than alignment loss, and (2) both losses are upper-bounded by some small value. Our proposed loss can be seen as learning one task through the scope of the other task. As the model is composed of a shared network and task-specific networks, cross-task consistency loss forces the model to efficiently pass the information of one task to the other and hold consistency between its predictions. Furthermore, both cross-task consistency loss and alignment loss are upper-bounded by $\xi_Y$, which tells us that cross-task terms are competitive with direct predictions.

\section{XTasC-Net} \label{sec:xtascnet}
\subsection{Model Architecture}
\begin{figure*}[!t]
\centering
\includegraphics[width=6.5in]{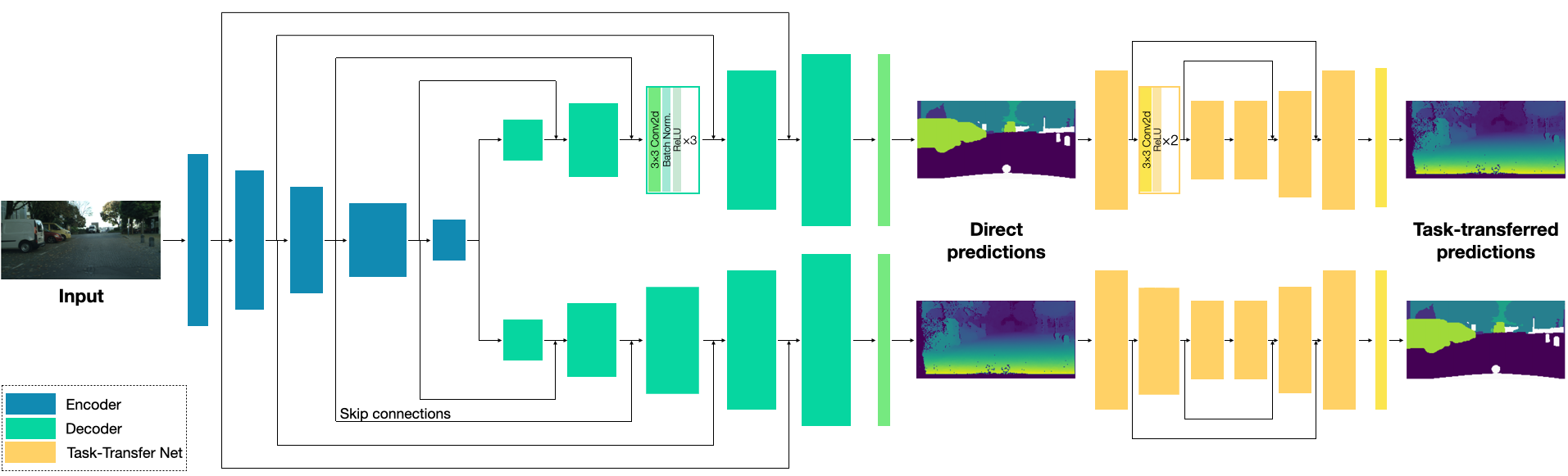}
\caption{Architecture overview of XTasC-Net.}
\label{fig:xtascnet}
\end{figure*}

Based on our cross-task consistency learning framework, we propose an original neural network model named \emph{XTasC-Net} (Cross-Task Consistency Network) to conduct the experiments. Fig. \ref{fig:xtascnet} shows our model which is built from 3 modules.

XTasC-Net is an encoder-decoder architecture with separate, individual decoders for each task. The encoder and decoder module is connected also with skip-connections, similar to U-Net \cite{ronne15}. The output of each decoder, the direct predictions, are then fed into separate networks, which we refer as \emph{TTNet} (Task-Transfer Networks), similar to \cite{zamir20}. The outputs of TTNet are the task-transferred predictions.

We choose ResNet \cite{he16}, a commonly used network for image encoding, as the backbone of the encoder. There are 5 types of ResNet, namely ResNet18, ResNet34, ResNet50, ResNet101, ResNet152. After comparing results the results, we decided to use ResNet34 which resulted in best performance with the least number of parameters (see Appendix \ref{sec:appendixB}).

The decoder consists of 5 blocks and 1 convolutional layer. Based on U-Net, each decoder block takes in the concatenation of feature maps of the previous block and the respective encoder block. It then processes through 3 layers of convolutional network with kernel size $3\times3$ followed by batch normalization and ReLU activation function. The number of channels are all set to 128. At the end of each block, the feature maps are upsampled by 2. The output of the 5th block is then fed into a $1\times1$ convolutional layer so that the output results in the required dimensions for a given task. For segmentation task, the number of classes to classify is the dimension, while for depth estimation, the dimension is 1.

TTNets are also designed similarly to U-Net with 3 blocks of contracting path and 3 blocks of expansive path. In the contracting path, each block consists of 2 repeating applications of $3\times3$ convolutional layer followed by ReLU activation function. At the end of each block, the feature maps are downsampled by a $2\times2$ max-pooling layer with stride 2. The number of channels are, in order, 64, 128, 256. On the other side, the expansive path is built symmetrically with 2 convolutional layers with the same kernel size followed by ReLU activation layer. At the end of each block, the feature maps are upsampled by 2. The output of the final block is then fed into a $1\times1$ convolutional layer similar to the decoder.

\subsection{Loss Functions}
As mentioned in section \ref{sec:method}, our proposed method is to add the cross-task consistency loss to the loss function. Below, we denote $y_1,y_2$ as the target value/class we wish to predict, $\hat{y}_1,\hat{y}_2$ as the direct predictions, and $\hat{y}_{2\rightarrow1},\hat{y}_{1\rightarrow2}$ as the task-transferred predictions for task 1 and 2 respectively. We prepare 2 types of loss functions: $\ell_1,\ell_2$ are the losses for direct predictions and $\ell_{2\rightarrow1},\ell_{1\rightarrow2}$ are the cross-task consistency losses. Using these notations, the loss of task $t$, $\mathcal{L}_t$, can be denoted as the weighted average of the 2 losses,
\begin{equation}
\mathcal{L}_t=((1-\lambda_t)\ell_t(\hat{y}_t,y_t)+ \lambda_t\ell_{t\rightarrow s}(\hat{y}_{t\rightarrow s},\hat{y}_t))
\end{equation}
where $\lambda_t\in[0,1]$ is the weight and $s$ is the other task.

Let task 1 be the segmentation task and task 2 be the depth estimation task.

Following the example of numerous previous works, we use cross-entropy loss as the loss function for the segmentation task. Given an $H\times W$ output, the cross entropy loss for direct predictions is expressed as,
\begin{equation}
    \ell_1(\hat{y}_1,y_1)=-\frac{1}{HW}\sum_{i=0}^{H-1}\sum_{j=0}^{W-1}y_1(i,j)\log\hat{y}_1(i,j)
\end{equation}
For task-transferred predictions, the loss is expressed as,
\begin{equation}
    \ell_{2\rightarrow1}(\hat{y}_{2\rightarrow1},\hat{y}_1)=
    -\frac{1}{HW}\sum_{i=0}^{H-1}\sum_{j=0}^{W-1}\hat{y}_1(i,j)\log\hat{y}_{2\rightarrow1}(i,j)
\end{equation}
We do not backpropagate the loss via direct prediction for cross-task consistency loss because the target we are trying to minimize against is the direct prediction, the output of the other task's decoder.

For depth estimation task, we use $l_1$ norm distance as the loss function. Given an $H\times W$ output, the depth loss for direct predictions is written as,
\begin{equation}
    \ell_2(\hat{y}_2,y_2)=\frac{1}{HW}\sum_{i=0}^{H-1}\sum_{j=0}^{W-1}|\hat{y}_2(i,j)-y_2(i,j)|
\end{equation}
for all pixels with valid depth values. During training, we mask the invalid pixels so that its loss is not backpropagated through the network. We also use $l_1$ norm for task-transferred predictions as well.
\begin{equation}
    \ell_{1\rightarrow2}(\hat{y}_{1\rightarrow2},\hat{y}_2)
    =\frac{1}{HW}\sum_{i=0}^{H-1}\sum_{j=0}^{W-1}|\hat{y}_{1\rightarrow2}(i,j)-\hat{y}_2(i,j)|
\end{equation}
Similar to task-transferred prediction of segmentation task, the loss is not backpropagated via the direct prediction.

Overall, the loss $\mathcal{L}_{\rm TOTAL}$ we wish to minimize is the weighted sum of the 2 tasks,
\begin{equation}
    \mathcal{L}_{\rm TOTAL}=\sum_{t=1}^2\omega_t\mathcal{L}_t
\end{equation}
where $\omega_1,\omega_2$ are the weights.  For $\omega_1,\omega_2$, we experiment using equal weights, uncertainty weights \cite{kendall18}, and GradNorm \cite{chen18}.

\section{Experiments}\label{sec:experiments}
\subsection{Datasets}
We consider 2 datasets, Cityscapes \cite{cordts16} and NYU \cite{silberman12} dataset to validate our proposed method.

Cityscapes dataset is a collection of diverse urban street scenes gathered using a stereo camera. It is provided with 19-class and 7-class segmentation labels, and we use the 7-class version so that it can be compared with previous works. The images' resolution is $1024\times2048$ but we resize to $128\times256$ to speed up training.

NYU dataset is a dataset composed of a wide variety of indoor scenes recorded by an RGB camera and depth cameras using Microsoft Kinect with 13-class segmentation labels. The resolution of the images are $480\times640$ but we resize to $288\times384$ to speed up training.

For both datasets, we apply normalization, random horizontal flipped with a probability of 0.5, and random scaled cropping with scales chosen randomly from $[1.0,1.2,1.5]$. We mask out invalid pixels during training, such as void classes of segmentation task and pixels with depth 0 (incorrectly calculated depth) for depth estimation task. Following previous works, for depth of Cityscapes, we use the inverse disparity values as the target because the raw disparity values range from 0 to infinity (ex. sky) and training to predict such infinite values lead to poor generalization.

\subsection{Evaluation Metrics} \label{sec:metrics}
We use the following metrics to evaluate the performance. For segmentation task, we use mean Intersection-over-Union (mIoU) and pixel accuracy (Pix. Acc.). For depth estimation task, we use absolute error (Abs. Err.) and absolute relative error (Rel. Err.).

Furthermore, following \cite{vande20,maninis19}, we compute performance improvement $\Delta_m$ of a model $m$ against the baseline model $b$ as the average percentage points' gain of $|M|$ evaluation metrics,
\begin{equation}
    \Delta_m=\frac{1}{|M|}\sum_{i\in[M]}(-1)^{l_i}\frac{M_{m,i}-M_{b,i}}{M_{b,i}}
\end{equation}
where $l_i=1$ if a lower value means better performance for measure $M_i$ of task $i$ and 0 otherwise.

\subsection{Training Protocols} \label{sec:protocol}
For all datasets, we use Adam optimizer with an initial learning rate of 0.0001. For Cityscapes dataset, the learning rate is halved every 80 epochs during training. We train the model for 250 epochs with batch size 8. For NYU dataset, we halve the learning rate every 60 epochs during training. We train the model for 100 epochs with batch size 6. We choose $\lambda_1,\lambda_2$ as 0.01 for Cityscapes and 0.0001 for NYU dataset.

We conduct all experiments using Pytorch. Our codes are available at \url{https://github.com/akarimoon/xtask_mt}.

\subsection{Baseline Models}
We compare the result of our XTasC-Net with 3 models, (1) the same encoder and decoder to train each task separately, (2) the same architecture as XTasC-Net but but without TTNet (by setting $\lambda_1=\lambda_2=0$), and (3) the same architecture as XTasC-Net but using alignment loss. We refer to these models as Base ST-Net, Base MT-Net, and Align-Net, respectively. For models except Base ST-Net, we use uncertainty weights \cite{kendall18} to balance the 2 tasks' loss when learning jointly. Further analysis on weighting methods is written in the appendix (Appendix \ref{sec:appendixC}).

\subsection{Results} \label{sec:result}
\subsubsection{Results on Cityscapes dataset}  \label{sec:cs_result}
First, we compare our results against the 3 baseline models with several weighting methods in Table \ref{table:cs_1}.

\begin{table*}[!t]
  \renewcommand{\arraystretch}{1.3}
  \caption{Ablation Results of Different Models on Cityscapes Validation Set for 7-Class Semantic Segmentation and Depth Estimation Task. Best Results are in \textbf{Bold}.
  }
  \label{table:cs_1}
  \centering
  \begin{tabular}{|c||c|c|c|c|c|}
    \hline
    \multicolumn{1}{|c||}{\multirow{3}{*}{Model}} &
    \multicolumn{2}{c|}{Segmentation} & \multicolumn{2}{c|}{Depth} & Performance \\ 
    & \multicolumn{2}{c|}{(Higher Better)} & \multicolumn{2}{c|}{(Lower Better)} & (Higher Better) \\ \cline{2-6}
    & mIoU & Pix Acc & Abs Err & Rel Err & $\Delta_m$ \\ \hline \hline
    Base ST-Net & 66.40 & 93.48 & 0.0124 & 19.78 & 0.00 \\ \hline
    Base MT-Net & \textbf{66.84} & \textbf{93.57} & 0.0122 & 19.61 & 1.44 \\ \hline
    Align-Net   & 66.61 & 93.53 & 0.0124 & 19.83 & 0.67 \\ \hline
    XTasC-Net   & 66.51 & 93.56 & \textbf{0.0122} & \textbf{19.40} & \textbf{1.58} \\ \hline
  \end{tabular}
\end{table*}

Overall, we can observe that our XTasC-Net succeeds in achieving higher results compared to the other 3 baseline models. Although Base MT-Net achieves the highest score for segmentation tasks, using TTNet to transfer losses between tasks leads to better overall performance. Between Align-Net and XTasC-Net, XTasC-Net achieves higher overall performance. These results show that task-transferred predictions do not interfere with direct predictions but exploit the task relationships to improve both tasks' predictions.

Table \ref{table:cs_2} shows the results for all methods.

\begin{table*}[!t]
  \renewcommand{\arraystretch}{1.3}
  \caption{Results of Multi-Task Learning on Cityscapes Validation Set for 7-Class Semantic Segmentation and Depth Estimation Task. \#P Shows the Number of Parameters of the Model.  \emph{Italic} Represents Estimated Values. 
  Best Results are in \textbf{Bold} and Second Best are \underline{Underlined}.
  * Equal Weights. {\dag} Uncertainty Weights. {\ddag} Gradient-based Weight Learning.
  }
  \label{table:cs_2}
  \centering
  \begin{tabular}{|l|c||c|c|c|c|}
    \hline
    \multicolumn{1}{|c|}{\multirow{3}{*}{Model}} & \multirow{3}{*}{\#P. $\left[\times10^7\right]$} & 
    \multicolumn{2}{c|}{Segmentation} & \multicolumn{2}{c|}{Depth} \\
    & & \multicolumn{2}{c|}{(Higher Better)} & \multicolumn{2}{c|}{(Lower Better)} \\ \cline{3-6}
    & & mIoU & Pix Acc & Abs Err & Rel Err \\ \hline \hline
    STAN \cite{liu19} & 12.52 & 51.90 & 90.87 & 0.0145 & 27.46 \\ \hline
    Dense{\dag} \cite{huang17} & 14.96 & 51.89 & 91.22 & 0.0134 & 25.36 \\ \hline
    Cross-Stitch{\dag} \cite{misra16} & $\approx8.24$ & 50.31 & 90.43 & 0.0152 & 31.36 \\ \hline
    MTAN* \cite{liu19} & 4.12 & 53.04 & 91.11 & 0.0144 & 33.63 \\ \hline
    PCGrad* \cite{yu20} & 4.12 & 53.59 & 91.45 & 0.0171 & 31.34 \\ \hline
    KD4MTL* \cite{li20} & 4.12 & 52.71 & 91.54 & 0.0139 & 27.33 \\ \hline
    AdaMT-Net{\ddag} \cite{jha20} & \textsl{4.91} & \underline{62.53} & \textbf{94.16} & \underline{0.0125} & \underline{22.23} \\ \hline
    XTasC-Net{\dag} (Ours) & 3.15 & \textbf{66.51} & \underline{93.56} & \textbf{0.0122} & \textbf{19.40} \\ \hline
  \end{tabular}
\end{table*}

We show the number of parameters in the table to show that our XTasC-Net achieves the best result with fewer parameters. The results show that our model outperforms previous works on most  evaluation metrics with a sufficiently fewer amount of parameters.

For segmentation task, we observe great improvement on mIoU metric. We think that this is due to the difference of using attention modules or not. Previous works such as \cite{liu19,jha20,li20,yu20} have used attention modules in their networks, enabling the model to ``look'' at the entire image in the training phase. On the other hand, since our model only uses convolutional layers, the model can only learn from pixels nearby. Intuitively, this leads to higher mIoU while attention modules improve pixel accuracy.

The result shows that our method reduces error both in terms of absolute error and relative error for the depth estimation task. Intuitively, learning from the segmentation task's predictions motivates the model to output different depth ranges for each class. We think that this leads to depth predictions with more contrast and hence higher accuracy.

\begin{figure*}[!t]
\centering
\includegraphics[width=6.5in]{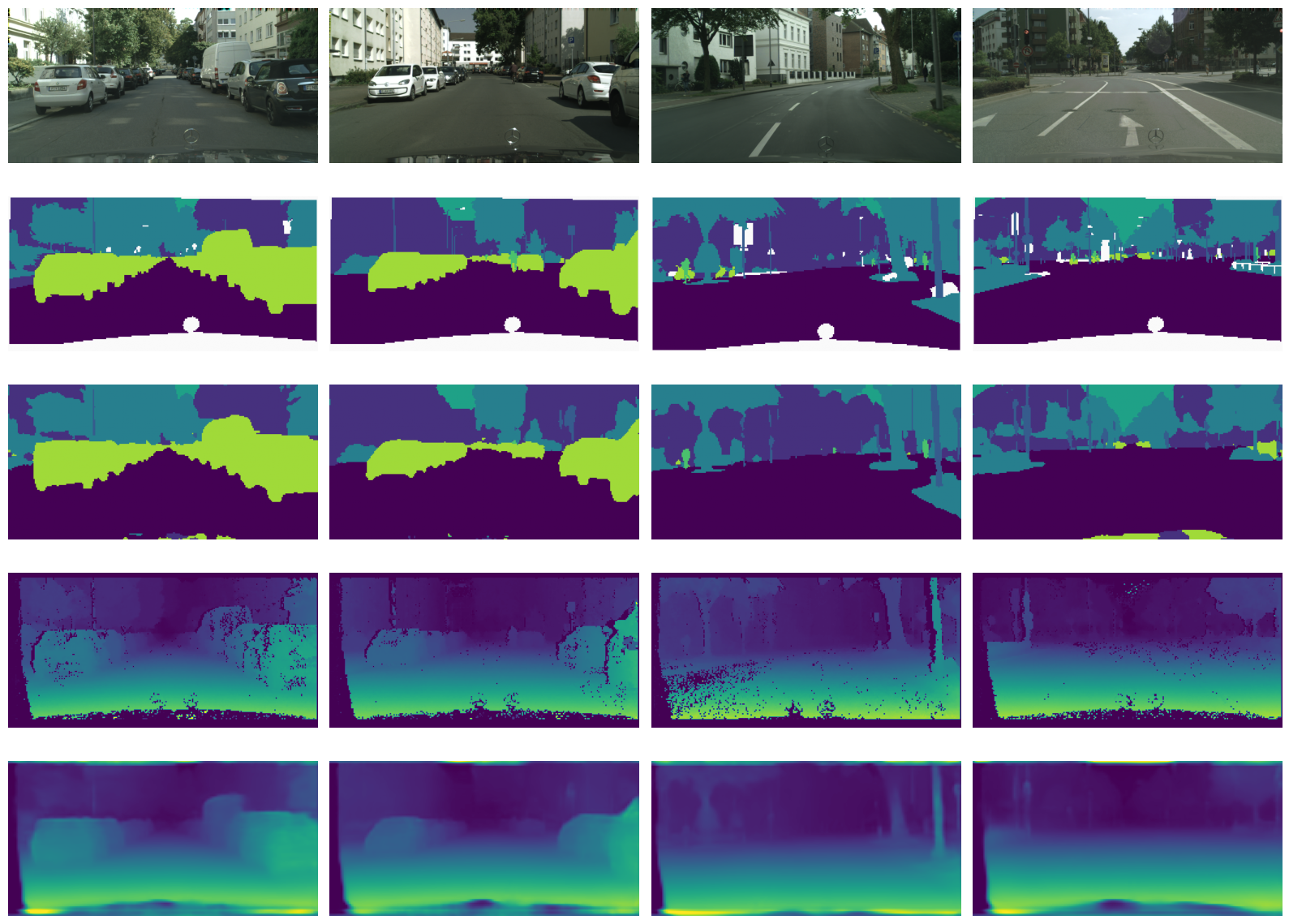}
\caption{Qualitative results on Cityscapes validation set. From top to bottom: input image, ground truth segmentation, predicted segmentation, ground truth depth, and predicted depth.}
\label{fig:cs_batch}
\end{figure*}

Our qualitative results are shown in Fig. \ref{fig:cs_batch}.

\subsubsection{Results on NYU dataset} \label{sec:nyu_result}
We compare our results against the 3 baseline models in Table \ref{table:nyu_1}.

\begin{table*}[!t]
  \renewcommand{\arraystretch}{1.3}
  \caption{Ablation Results of Different Models on NYU Test Set for 13-Class Semantic Segmentation and Depth Estimation Task. Best Results are in \textbf{Bold}.
  }
  \label{table:nyu_1}
  \centering
  \begin{tabular}{|c||c|c|c|c|c|}
    \hline
    \multicolumn{1}{|c||}{\multirow{3}{*}{Model}} &
    \multicolumn{2}{c|}{Segmentation} & \multicolumn{2}{c|}{Depth} & Performance \\ 
    & \multicolumn{2}{c|}{(Higher Better)} & \multicolumn{2}{c|}{(Lower Better)} & (Higher Better) \\ \cline{2-6}
    & mIoU & Pix Acc & Abs Err & Rel Err & $\Delta_m$ \\ \hline \hline
    Base ST-Net & 30.62 & 62.30 & 0.6451 & 0.2443 & 0.00 \\ \hline
    Base MT-Net & \textbf{31.02} & \textbf{63.58} & 0.6103 & 0.2284 & 3.82 \\ \hline
    Align-Net   & 30.46 & 63.32 & 0.6514 & 0.2295 & 1.55 \\ \hline
    XTasC-Net.  & 30.31 & 63.02 & \textbf{0.5954} & \textbf{0.2235} & \textbf{4.09} \\ \hline
  \end{tabular}
\end{table*}

Again, we see a slight drop of performance for segmentation task. Compared to Cityscapes dataset, we observe that using MTL results in lower mIoU score because scenes in NYU datasets differ largely, with different depth range, lighting condition, and camera angle. However, using the proposed methods lead to better performance.

Table \ref{table:nyu_2} shows the result for all methods. 

\begin{table*}[!t]
  \renewcommand{\arraystretch}{1.3}
  \caption{Results of Multi-Task Learning on NYU Test Set for 13-Class Semantic Segmentation and Depth Estimation Task.  Best Results are in \textbf{Bold} and Second Best are \underline{Underlined}.
  * Equal weights. {\dag} Uncertainty weights. {\ddag} Gradient-based weight learning. $^\star$ DWA.
  }
  \label{table:nyu_2}
  \centering
  \begin{tabular}{|l||c|c|c|c|}
    \hline
    \multicolumn{1}{|c||}{\multirow{3}{*}{Model}} & \multicolumn{2}{c|}{Segmentation} & \multicolumn{2}{c|}{Depth} \\ 
    & \multicolumn{2}{c|}{(Higher Better)} & \multicolumn{2}{c|}{(Lower Better)}\\ \cline{2-5}
    & mIoU & Pix Acc & Abs Err & Rel Err \\ \hline \hline
    STAN \cite{liu19} & 16.65 & 55.07 & 0.6935 & 0.2891\\ \hline
    Dense \cite{huang17} & 17.22 & 55.59 & \underline{0.6002} & 0.2654 \\ \hline
    Cross-Stitch \cite{misra16} & 17.01 & 53.99 & 0.6095 & 0.2671 \\ \hline
    MTAN$^\star$ \cite{liu19} & 20.10 & 53.73 & 0.6417 & 0.2758 \\ \hline
    PCGrad{\dag} \cite{yu20} & 21.29 & 54.07 & 0.6705 & 0.3000 \\ \hline
    KD4MTL* \cite{li20} & \underline{22.44} & 57.32 & 0.6003 & 0.2601 \\ \hline
    AdaMT-Net{\ddag} \cite{jha20} & 20.61 & \underline{58.91} & 0.6136 & \underline{0.2547} \\ \hline
    XTasC-Net\dag (Ours) & \textbf{30.31} & \textbf{63.02} & \textbf{0.5954} & \textbf{0.2235} \\ \hline
  \end{tabular}
\end{table*}

As shown in the table, our model outperforms all previous works except for AdaMT-Net \cite{jha20}. Compared to AdaMT-Net, our model improves mIoU and relative depth error by a fair margin.

For the segmentation task, we again observe a larger improvement for mIoU compared to pixel accuracy. As described in section \ref{sec:cs_result}, we can infer that this is due to the models' characteristics. Using only convolutional layers without any attention modules, we achieve a 7.87 point improvement of mIoU and a smaller gain of 2.67 points for pixel accuracy.  Our model shows significant performance for depth estimation tasks as well (Abs Err: 0.6002 $\rightarrow$ 0.5954, Abs Rel Err: 0.2547 $\rightarrow$ 0.2235).

\section{Conclusion}\label{sec:conclusion}
In this work, we made mainly two contributions. First, we proposed a new model architecture called XTasC-Net (Cross-Task Consistency Network) that achieved state-of-the-art results for most evaluation metrics on two benchmark datasets, Cityscapes and NYU dataset. We also showed that our model is parameter-efficient. Secondly, we introduced a new loss term, cross-task consistency loss, for the 2-task MTL setting. Our method performs well on both datasets that were collected using a stereo camera (Cityscapes dataset) and infrared camera (NYU dataset).

Cross-task consistency loss is a term that takes the loss between the transferred prediction with the prediction of the other task. This loss motivated the model to output consistent predictions for both tasks by transferring information of one task to the other. We showed both theoretically and empirically the effect of cross-task consistency loss. Although our proposed loss term is limited to 2-task MTL problems, it efficiently uses the task relationship for performance improvement. Furthermore, compared to previous self-supervised learning methods, our self-supervised training method can be applied even if the data were collected using non-stereo cameras.

While we only explored the case of semantic segmentation and depth estimation task, we expect our proposed framework can be applied to any MTL settings if there exists some relationship between the tasks. In the field of computer vision, many tasks are related to each other \cite{zamir18}. With the introduction of datasets aimed for MTL \cite{zamir18,roberts20}, we think that our framework will help the model learn and exploit the task relationships. Furthermore, if the framework can be extended to an arbitrary number of tasks, it will help our understanding of task relationships.



\ifCLASSOPTIONcaptionsoff
  \newpage
\fi



\bibliographystyle{IEEEtran}
\bibliography{IEEEabrv,references.bib}

\begin{thebibliography}{10}
\providecommand{\url}[1]{#1}
\csname url@samestyle\endcsname
\providecommand{\newblock}{\relax}
\providecommand{\bibinfo}[2]{#2}
\providecommand{\BIBentrySTDinterwordspacing}{\spaceskip=0pt\relax}
\providecommand{\BIBentryALTinterwordstretchfactor}{4}
\providecommand{\BIBentryALTinterwordspacing}{\spaceskip=\fontdimen2\font plus
\BIBentryALTinterwordstretchfactor\fontdimen3\font minus
  \fontdimen4\font\relax}
\providecommand{\BIBforeignlanguage}[2]{{%
\expandafter\ifx\csname l@#1\endcsname\relax
\typeout{** WARNING: IEEEtran.bst: No hyphenation pattern has been}%
\typeout{** loaded for the language `#1'. Using the pattern for}%
\typeout{** the default language instead.}%
\else
\language=\csname l@#1\endcsname
\fi
#2}}
\providecommand{\BIBdecl}{\relax}
\BIBdecl

\bibitem{long17}
M.~Long, Z.~Cao, J.~Wang, and P.~S. Yu, ``Learning multiple tasks with
  multilinear relationship networks,'' in \emph{Proc. Adv. Neural Inf. Process.
  Syst. (NIPS)}, vol.~30, 2017, pp. 1594--1603.

\bibitem{misra16}
I.~{Misra}, A.~{Shrivastava}, A.~{Gupta}, and M.~{Hebert}, ``Cross-stitch
  networks for multi-task learning,'' in \emph{Proc. IEEE/CVF Conf. Comput.
  Vis. Pattern Recognit. (CVPR)}, 2016, pp. 3994--4003.

\bibitem{yang19}
P.~Yang, Q.~Tan, J.~Ye, H.~Tong, and J.~He, ``Deep multi-task learning with
  adversarial-and-cooperative nets,'' in \emph{Proc. 28th Int. Joint Conf.
  Artif. Intell. (IJCAI)}, 7 2019, pp. 4078--4084.

\bibitem{liu19}
S.~{Liu}, E.~{Johns}, and A.~J. {Davison}, ``End-to-end multi-task learning
  with attention,'' in \emph{Proc. IEEE/CVF Conf. Comput. Vis. Pattern
  Recognit. (CVPR)}, 2019, pp. 1871--1880.

\bibitem{jha20}
A.~{Jha}, A.~{Kumar}, B.~{Banerjee}, and S.~{Chaudhuri}, ``Adamt-net: An
  adaptive weight learning based multi-task learning model for scene
  understanding,'' in \emph{Proc. IEEE/CVF Conf. Comput. Vis. Pattern Recognit.
  Workshops (CVPRW)}, 2020, pp. 3027--3035.

\bibitem{kendall18}
A.~Kendall, Y.~Gal, and R.~Cipolla, ``Multi-task learning using uncertainty to
  weigh losses for scene geometry and semantics,'' in \emph{Proc. IEEE/CVF
  Conf. Comput. Vis. Pattern Recognit. (CVPR)}, 2018, pp. 7482--7491.

\bibitem{chen18}
Z.~Chen, V.~Badrinarayanan, C.-Y. Lee, and A.~Rabinovich, ``{G}rad{N}orm:
  Gradient normalization for adaptive loss balancing in deep multitask
  networks,'' in \emph{Proc. 35th Int. Conf. Mach. Learn. (ICML)}, 2018, pp.
  794--803.

\bibitem{yu20}
T.~Yu, S.~Kumar, A.~Gupta, S.~Levine, K.~Hausman, and C.~Finn, ``Gradient
  surgery for multi-task learning,'' in \emph{Advances in Neural Information
  Processing Systems}, 2020, pp. 5824--5836.

\bibitem{chen19}
P.~{Chen}, A.~H. {Liu}, Y.~{Liu}, and Y.~F. {Wang}, ``Towards scene
  understanding: Unsupervised monocular depth estimation with semantic-aware
  representation,'' in \emph{Proc. IEEE/CVF Conf. Comput. Vis. Pattern
  Recognit. (CVPR)}, 2019, pp. 2619--2627.

\bibitem{klingner20}
M.~Klingner, J.-A. Termöhlen, J.~Mikolajczyk, and T.~Fingscheidt,
  ``Self-supervised monocular depth estimation: Solving the dynamic object
  problem by semantic guidance,'' in \emph{Proc. Eur. Conf. Comput. Vis.
  (ECCV)}.\hskip 1em plus 0.5em minus 0.4em\relax Cham: Springer, 2020, pp.
  582--600.

\bibitem{silberman12}
N.~Silberman, D.~Hoiem, P.~Kohli, and R.~Fergus, ``Indoor segmentation and
  support inference from rgbd images,'' in \emph{Proc. Eur. Conf. Comput. Vis.
  (ECCV)}.\hskip 1em plus 0.5em minus 0.4em\relax Cham: Springer, 2012, pp.
  746--760.

\bibitem{zhu17}
J.~{Zhu}, T.~{Park}, P.~{Isola}, and A.~A. {Efros}, ``Unpaired image-to-image
  translation using cycle-consistent adversarial networks,'' in \emph{Proc.
  Int. Conf. Comput. Vis. (ICCV)}, 2017, pp. 2242--2251.

\bibitem{chen20}
T.~Chen, S.~Kornblith, M.~Norouzi, and G.~Hinton, ``A simple framework for
  contrastive learning of visual representations,'' in \emph{Proc. 37th Int.
  Conf. Mach. Learn. (ICML)}, 2020, pp. 1597--1607.

\bibitem{tosh20}
\BIBentryALTinterwordspacing
C.~Tosh, A.~Krishnamurthy, and D.~Hsu, ``Contrastive learning, multi-view
  redundancy, and linear models,'' 2020. [Online]. Available:
  \url{http://arxiv.org/abs/2008.10150}
\BIBentrySTDinterwordspacing

\bibitem{burge10}
J.~Burge, C.~C. Fowlkes, and M.~S. Banks, ``Natural-scene statistics predict
  how the figure–ground cue of convexity affects human depth perception,'' in
  \emph{J. Neuroscience}, vol.~30, 2010, pp. 7269--7280.

\bibitem{cordts16}
M.~Cordts, M.~Omran, S.~Ramos, T.~Rehfeld, M.~Enzweiler, R.~Benenson,
  U.~Franke, S.~Roth, and B.~Schiele, ``The cityscapes dataset for semantic
  urban scene understanding,'' in \emph{Proc. IEEE/CVF Conf. Comput. Vis.
  Pattern Recognit. (CVPR)}, 2016, pp. 3213--3223.

\bibitem{caruana98}
R.~Caruana, ``Multitask learning,'' in \emph{Learning to Learn}, S.~Thrun and
  L.~Pratt, Eds.\hskip 1em plus 0.5em minus 0.4em\relax Boston, MA: Springer,
  1998, ch.~5, pp. 95--133.

\bibitem{ruder17}
\BIBentryALTinterwordspacing
S.~Ruder, ``An overview of multi-task learning in deep neural networks,'' 2017.
  [Online]. Available: \url{http://arxiv.org/abs/1706.05098}
\BIBentrySTDinterwordspacing

\bibitem{micchelli04}
C.~A. Micchelli and M.~Pontil, ``Kernels for multi–task learning,'' in
  \emph{Proc. Adv. Neural Inf. Process. Syst. (NIPS)}, 2004, pp. 921--928.

\bibitem{ciliberto10}
C.~Ciliberto, Y.~Mroueh, T.~Poggio, and L.~Rosasco, ``Convex learning of
  multiple tasks and their structure,'' in \emph{Proc. 32nd Int. Conf. Mach.
  Learn. (ICML)}, 2015, p. 1548–1557.

\bibitem{zhang10}
Y.~Zhang and D.-Y. Yeung, ``A convex formulation for learning task
  relationships in multi-task learning,'' in \emph{Proc. 26th Conf. Uncert.
  Artif. Intell. (UAI)}, 2010, p. 733–742.

\bibitem{ciliberto17}
C.~Ciliberto, A.~Rudi, L.~Rosasco, and M.~Pontil, ``Consistent multitask
  learning with nonlinear output relations,'' in \emph{Proc. Adv. Neural Inf.
  Process. Syst. (NIPS)}, vol.~30, 2017, pp. 1986--1996.

\bibitem{zamir18}
A.~R. {Zamir}, A.~{Sax}, W.~{Shen}, L.~{Guibas}, J.~{Malik}, and S.~{Savarese},
  ``Taskonomy: Disentangling task transfer learning,'' in \emph{Proc. IEEE/CVF
  Conf. Comput. Vis. Pattern Recognit. (CVPR)}, 2018, pp. 3712--3722.

\bibitem{zhou21}
F.~{Zhou}, C.~{Shui}, M.~{Abbasi}, L.~E. {Robitaille}, B.~{Wang}, and
  C.~{Gagné}, ``Task similarity estimation through adversarial multitask
  neural network,'' \emph{IEEE Transactions on Neural Networks and Learning
  Systems}, vol.~32, no.~2, pp. 466--480, 2021.

\bibitem{hinton15}
\BIBentryALTinterwordspacing
G.~Hinton, O.~Vinyals, and J.~Dean, ``Distilling the knowledge in a neural
  network,'' in \emph{NIPS Deep Learning and Representation Learning Workshop},
  2015. [Online]. Available: \url{http://arxiv.org/abs/1503.02531}
\BIBentrySTDinterwordspacing

\bibitem{xu18}
D.~{Xu}, W.~{Ouyang}, X.~{Wang}, and N.~{Sebe}, ``Pad-net: Multi-tasks guided
  prediction-and-distillation network for simultaneous depth estimation and
  scene parsing,'' in \emph{Proc. IEEE/CVF Conf. Comput. Vis. Pattern Recognit.
  (CVPR)}, 2018, pp. 675--684.

\bibitem{vande19}
\BIBentryALTinterwordspacing
S.~Vandenhende, S.~Georgoulis, B.~D. Brabandere, and L.~{Van Gool}, ``Branched
  multi-task networks: Deciding what layers to share,'' 2019. [Online].
  Available: \url{http://arxiv.org/abs/1904.02920}
\BIBentrySTDinterwordspacing

\bibitem{zhang19}
Z.~Zhang, Z.~Cui, C.~Xu, Y.~Yan, N.~Sebe, and J.~Yang, ``Pattern-affinitive
  propagation across depth, surface normal and semantic segmentation,'' in
  \emph{Proc. IEEE/CVF Conf. Comput. Vis. Pattern Recognit. (CVPR)}, 2019, pp.
  4106--4115.

\bibitem{li20}
W.-H. Li and H.~Bilen, ``Knowledge distillation for multi-task learning,'' in
  \emph{Proc. Eur. Conf. Comput. Vis. Workshops (ECCVW)}.\hskip 1em plus 0.5em
  minus 0.4em\relax Cham: Springer, 2020, pp. 163--176.

\bibitem{vande20}
S.~Vandenhende, S.~Georgoulis, and L.~{Van Gool}, ``Mti-net: Multi-scale task
  interaction networks for multi-task learning,'' in \emph{Proc. Eur. Conf.
  Comput. Vis. (ECCV)}.\hskip 1em plus 0.5em minus 0.4em\relax Cham: Springer,
  2020, pp. 527--543.

\bibitem{mousa16}
A.~{Mousavian}, H.~{Pirsiavash}, and J.~{Košecká}, ``Joint semantic
  segmentation and depth estimation with deep convolutional networks,'' in
  \emph{Proc. 4th Int. Conf. 3D Vis.}, 2016, pp. 611--619.

\bibitem{zhang18}
Z.~Zhang, Z.~Cui, C.~Xu, Z.~Jie, X.~Li, and J.~Yang, ``Joint task-recursive
  learning for semantic segmentation and depth estimation,'' in \emph{Proc.
  Eur. Conf. Comput. Vis. (ECCV)}, September 2018, pp. 235--251.

\bibitem{sener18}
O.~Sener and V.~Koltun, ``Multi-task learning as multi-objective
  optimization,'' in \emph{Proc. Adv. Neural Inf. Process. Syst. (NeurIPS)},
  2018, pp. 525--536.

\bibitem{godard17}
C.~Godard, O.~{Mac Aodha}, and G.~J. Brostow, ``Unsupervised monocular depth
  estimation with left-right consistency,'' in \emph{Proc. IEEE/CVF Conf.
  Comput. Vis. Pattern Recognit. (CVPR)}, 2017.

\bibitem{godard19}
C.~Godard, O.~{Mac Aodha}, M.~Firman, and G.~J. Brostow, ``Digging into
  self-supervised monocular depth prediction,'' in \emph{Proc. Int. Conf.
  Comput. Vis. (ICCV)}, 2019.

\bibitem{tiao18}
\BIBentryALTinterwordspacing
L.~C. Tiao, E.~V. Bonilla, and F.~Ramos, ``Cycle-consistent adversarial
  learning as approximate bayesian inference,'' 2018. [Online]. Available:
  \url{http://arxiv.org/abs/1806.01771}
\BIBentrySTDinterwordspacing

\bibitem{ronne15}
O.~Ronneberger, P.~Fischer, and T.~Brox, ``U-net: Convolutional networks for
  biomedical image segmentation,'' \emph{Med. Image Comput. Comput.-Assisted
  Intervention (MICCAI)}, pp. 234--241, 2015.

\bibitem{zamir20}
A.~R. {Zamir}, A.~{Sax}, N.~{Cheerla}, R.~{Suri}, Z.~{Cao}, J.~{Malik}, and
  L.~J. {Guibas}, ``Robust learning through cross-task consistency,'' in
  \emph{Proc. IEEE/CVF Conf. Comput. Vis. Pattern Recognit. (CVPR)}, 2020, pp.
  11\,194--11\,203.

\bibitem{he16}
K.~{He}, X.~{Zhang}, S.~{Ren}, and J.~{Sun}, ``Deep residual learning for image
  recognition,'' in \emph{Proc. IEEE/CVF Conf. Comput. Vis. Pattern Recognit.
  (CVPR)}, 2016, pp. 770--778.

\bibitem{maninis19}
K.~K. Maninis, I.~Radosavovic, and I.~Kokkinos, ``Attentive single-tasking of
  multiple tasks,'' in \emph{Proc. IEEE/CVF Conf. Comput. Vis. Pattern
  Recognit. (CVPR)}, 2019, p. 1851–1860.

\bibitem{huang17}
G.~Huang, Z.~Liu, L.~{Van Der Maaten}, and K.~Q. Weinberger, ``Densely
  connected convolutional networks,'' in \emph{Proc. IEEE/CVF Conf. Comput.
  Vis. Pattern Recognit. (CVPR)}, 2017, p. 4700–4708.

\bibitem{roberts20}
\BIBentryALTinterwordspacing
M.~Roberts and N.~Paczan, ``Hypersim: A photorealistic synthetic dataset for
  holistic indoor scene understanding,'' 2020. [Online]. Available:
  \url{http://arxiv.org/abs/2011.02523}
\BIBentrySTDinterwordspacing

\end{thebibliography}

\appendices
\section{Proof of the Proposition} \label{sec:appendixA}
\begin{IEEEproof}
By the law of total expectation, the first part of the equation is,
\setlength{\arraycolsep}{0.0em}
\begin{eqnarray*}
&\mathbb{E}&[(\mathbb{E}[\mathbb{E}[Y|X]|\mathbb{E}[Z|X]]-\mathbb{E}[Y|X])^2]\\
&&{=}\:\mathbb{E}[(\mathbb{E}[Y|X]-\mathbb{E}[Y|X])^2]\\
&&{=}\:0
\end{eqnarray*}
\setlength{\arraycolsep}{5pt}
The first inequality of the equation holds because,
\setlength{\arraycolsep}{0.0em}
\begin{eqnarray*}
&\mathbb{E}&[(\mathbb{E}[Y|\mathbb{E}[Z|X]]-\mathbb{E}[Y|X])^2]\\
&&{=}\mathbb{E}[(\mathbb{E}[Y|\mathbb{E}[Z|X]]-\mathbb{E}[\mathbb{E}[Y|X]|\mathbb{E}[Z|X]])^2]\\
&&{=}\mathbb{E}[(\mathbb{E}[Y-\mathbb{E}[Y|X]|\mathbb{E}[Z|X]])^2]\\
&&{\geq}\:0
\end{eqnarray*}
\setlength{\arraycolsep}{5pt}
Furthermore, the second inequality holds from Jensen's inequality,
\setlength{\arraycolsep}{0.0em}
\begin{eqnarray*}
&\mathbb{E}&[(\mathbb{E}[Y|\mathbb{E}[Z|X]]-\mathbb{E}[Y|X])^2]\\
&&{=}\mathbb{E}[(\mathbb{E}[Y-\mathbb{E}[Y|X]|\mathbb{E}[Z|X]])^2]\\
&&{\leq}\mathbb{E}[\mathbb{E}[(Y-\mathbb{E}[Y|X])^2|\mathbb{E}[Z|X]]]\\
&&{=}\mathbb{E}[(Y-\mathbb{E}[Y|X])^2]=\xi_Y
\end{eqnarray*}
\setlength{\arraycolsep}{5pt}
\end{IEEEproof}

\section{Effects of Different Encoders} \label{sec:appendixB}
As explained in section \ref{sec:xtascnet}, we use ResNet as our encoder. ResNet has several variations of depth, and different works have used different encoders as their encoder. Therefore, we also evaluate the change of performance between 3 types of ResNet, ResNet18, 34, and 50 (Table \ref{table:ablation_cs}, \ref{table:ablation_nyu}). We evaluate ResNet18 because it has fewer parameters compared to ResNet34. We also evaluate ResNet50 because it has more parameters but no more than the other models used for comparison.

\begin{table}[hbtp]
  \renewcommand{\arraystretch}{1.3}
  \caption{Ablation Results of Different Encoders on Cityscapes Validation Set for 7-class Semantic Segmentation and Depth Estimation Task. \#P Shows the Number of Parameters of the Model. * Use Weights Pretrained on ImageNet.}
  \label{table:ablation_cs}
  \centering
  \begin{tabular}{|c|c||c|c|c|c|}
    \hline
    \multicolumn{1}{|c|}{\multirow{3}{*}{ResNet}} & \multicolumn{1}{c||}{\multirow{3}{*}{\#P. $\left[\times10^7\right]$}} & 
    \multicolumn{2}{c|}{Segmentation} &  \multicolumn{2}{c|}{Depth} \\
    & & \multicolumn{2}{c|}{(Higher Better)} &  \multicolumn{2}{c|}{(Lower Better)} \\ \cline{3-6}
    & & mIoU & Pix Acc & Abs Err & Rel Err\\ \hline \hline
    18 & 2.14 & 66.15 & 93.47 & 0.0124 & 19.28 \\ \hline
    34 & 3.15 & 66.51 & 93.56 & 0.0122 & 19.40 \\ \hline
    34* & 3.15 & 68.71 & 94.23 & 0.0111 & 18.48 \\ \hline
    50 & 4.04 & 66.12 & 93.50 & 0.0124 & 19.98 \\ \hline
  \end{tabular}
\end{table}

\begin{table}[hbtp]
  \renewcommand{\arraystretch}{1.3}
  \caption{Ablation Results of Different Encoders on NYU Test Set for 13-class Semantic Segmentation and Depth Estimation Task. \#P Shows the Number of Parameters of the Model. * Use Weights Pretrained on ImageNet.}
  \label{table:ablation_nyu}
  \centering
  \begin{tabular}{|c|c||c|c|c|c|}
    \hline
    \multicolumn{1}{|c|}{\multirow{3}{*}{ResBet}} & \multicolumn{1}{c||}{\multirow{3}{*}{\#P. $\left[\times10^7\right]$}} & 
    \multicolumn{2}{c|}{Segmentation} &  \multicolumn{2}{c|}{Depth} \\ 
    & & \multicolumn{2}{c|}{(Higher Better)} &  \multicolumn{2}{c|}{(Lower Better)} \\ \cline{3-6}
    & & mIoU & Pix Acc & Abs Err & Rel Err \\ \hline \hline
    18 & 2.14 & 29.72 & 61.44 & 0.6174 & 0.2314 \\ \hline
    34 & 3.15 & 30.31 & 63.02 & 0.5954 & 0.2235 \\ \hline
    34* & 3.15 & 44.75 & 75.81 & 0.4851 & 0.1835 \\ \hline
    50 & 4.04 & 28.63 & 61.66 & 0.6115 & 0.2287 \\ \hline
  \end{tabular}
\end{table}

The results show that ResNet34 is the best encoder for both datasets resulting in the best scores for most evaluation metrics. Naturally, using a deeper encoder motivates the model to learn more contextual features and improve performance. However, for our model, the results show no such improvement by using deeper ResNet. The results also confirm that using pretrained weights of ResNet34 leads to better performance. Although the results in section \ref{sec:result} use the scores achieved without using pretrained weights for fair comparison, one should consider using the weights in applications.

\section{Effects of Different Weighting Methods} \label{sec:appendixC}
Below in table \ref{table:ablation2_cs} and \ref{table:ablation2_nyu}, we examine the effects of using different weighting methods. We consider using equal weights, uncertainty weights \cite{kendall18}, or gradient normalization \cite{chen18}.

\begin{table*}
  \caption{Ablation Results of Different Weighting Methods on Cityscapes Validation Set for 7-Class Semantic Segmentation and Depth Estimation Task.}
  \label{table:ablation2_cs}
  \centering
  \begin{tabular}{|c|l||c|c|c|c|c|}
    \hline
    \multicolumn{1}{|c|}{\multirow{3}{*}{Model}} & \multicolumn{1}{c||}{\multirow{3}{*}{Weighting}} & 
    \multicolumn{2}{c|}{Segmentation} & \multicolumn{2}{c|}{Depth} & Performance \\ 
    & & \multicolumn{2}{c|}{(Higher Better)} &  \multicolumn{2}{c|}{(Lower Better)} & (Higher Better) \\ \cline{3-7}
    & & mIoU & Pix Acc & Abs Err & Rel Err & $\Delta_m$ \\ \hline \hline
    \multirow{2}{*}{Base MT-Net} & Equal weights & 65.86 & 93.25 & 0.0129 & 20.47 & -1.50 \\ \cline{2-7}
    & Uncert. weights \cite{kendall18} & 66.84 & 93.57 & 0.0122 & 19.61 & 1.44 \\ \hline
    \multirow{3}{*}{XTasC-Net} & Equal weights & 66.32 & 93.51 & 0.0126 & 21.19 & -1.55 \\ \cline{2-7}
    & Uncert. weights \cite{kendall18} & 66.51 & 93.56 & 0.0122 & 19.40 & 1.58 \\ \cline{2-7}
    & GradNorm \cite{chen18} & 66.48 & 93.52 & 0.0124 & 19.58 & 0.93 \\ \hline 
  \end{tabular}
\end{table*}

\begin{table*}
  \caption{Ablation Results of Different Weighting Methods on NYU Test Set for 13-Class Semantic Segmentation and Depth Estimation Task. 
  }
  \label{table:ablation2_nyu}
  \centering
  \begin{tabular}{|c|l||c|c|c|c|c|}
    \hline
    \multicolumn{1}{|c|}{\multirow{3}{*}{Model}} & \multicolumn{1}{c||}{\multirow{3}{*}{Weighting}} & 
    \multicolumn{2}{c|}{Segmentation} &  \multicolumn{2}{c|}{Depth} & Performance \\
    & & \multicolumn{2}{c|}{(Higher Better)} &  \multicolumn{2}{c|}{(Lower Better)} & (Higher Better) \\ \cline{3-7}
    & & mIoU & Pix Acc & Abs Err & Rel Err & $\Delta_m$ \\ \hline \hline
    \multirow{2}{*}{Base MT-Net} & Equal weights & 30.42 & 62.89 & 0.6384 & 0.2326 & 1.53 \\ \cline{2-7}
    & Uncert. weights \cite{kendall18} & 31.02 & 64.58 & 0.6103 & 0.2284 & 3.82 \\ \hline
    \multirow{3}{*}{XTasC-Net} & Equal weights & 30.65 & 63.13 & 0.6319 & 0.2237 & 2.98 \\ \cline{2-7}
    & Uncert. weights \cite{kendall18} & 30.31 & 63.02 & 0.5954 & 0.2235 & 4.09 \\ \cline{2-7}
    & GradNorm \cite{chen18} & 30.71 & 63.44 & 0.6002 & 0.2222 & 4.53 \\ \hline 
  \end{tabular}
\end{table*}

Compared to Base MT-Net with equal weights, we observe that XTasC-Net with uncertainty weights or GradNorm leads to improvement in all four evaluation metrics except uncertainty weights for NYU dataset. Between the two weighting methods, we find that the model with uncertainty weighting excels especially for depth estimation task. Furthermore, under uncertainty weighting regime, XTasC-Net has better overall scores than Base MT-Net (Cityscapes: 1.44 $\rightarrow$ 1.58, NYU: 3.82 $\rightarrow$ 4.09).

\end{document}